\documentclass[10pt,twocolumn,letterpaper]{article}

\usepackage{cvpr}
\usepackage{times}
\usepackage{epsfig}
\usepackage{graphicx}
\usepackage{amsmath}
\usepackage{amssymb}

\usepackage{verbatim}
\usepackage{enumitem}

\usepackage{bbm, dsfont}

\usepackage[pagebackref=true,breaklinks=true,letterpaper=true,colorlinks,bookmarks=false]{hyperref}

\cvprfinalcopy 

\def\cvprPaperID{2623} 

\newcommand{\eat}[1]{} 

\eat{
\newcommand{\lu}[1]{{\textcolor{red}{[lu: #1]}}}
\newcommand{\nam}[1]{{\textcolor{blue}{[nam: #1]}}}
\newcommand{\cs}[1]{{\textcolor{green}{[chen: #1]}}}
\newcommand{\km}[1]{{\textcolor{orange}{[kevin: #1]}}}
}

\newcommand*{\ShowNotes}{} 

\ifdefined\ShowNotes
  \newcommand{\colornote}[3]{{\color{#1}\bf{#2: #3}\normalfont}}
\else
  \newcommand{\colornote}[3]{}
\fi

\eat{
\definecolor{darkred}{rgb}{0.7,0.1,0.1}
\definecolor{darkgreen}{rgb}{0.1,0.7,0.1}
\definecolor{cyan}{rgb}{0.7,0.0,0.7}
\definecolor{dblue}{rgb}{0.2,0.2,0.8}
\definecolor{maroon}{rgb}{0.76,.13,.28}
\definecolor{burntorange}{rgb}{0.81,.33,0}
\definecolor{tealblue}{rgb}{0.212,0.459, 0.533}
}

\newcommand {\lu}[1]{\colornote{red}{Lu}{#1}}
\newcommand {\nam}[1]{\colornote{green}{Nam}{#1}}
\newcommand {\km}[1]{\colornote{blue}{Kevin}{#1}}
\newcommand {\cs}[1]{\colornote{orange}{Chen}{#1}}

\renewcommand{\lu}[1]{}
\renewcommand{\nam}[1]{}
\renewcommand{\cs}[1]{}
\renewcommand{\km}[1]{}

\newcommand{\otherfancyname}{TIRG\xspace}
\newcommand{\otherfancynameLong}{Text Image Residual Gating\xspace}

\newcommand{\fquery}{f_{\mathrm{combine}}}
\newcommand{\fsim}{f_{\mathrm{sim}}}
\newcommand{\fimg}{f_{\mathrm{img}}}
\newcommand{\ftext}{f_{\mathrm{text}}}
\newcommand{\fgate}{f_{\mathrm{gate}}}
\newcommand{\fres}{f_{\mathrm{res}}}
\newcommand{\MLP}{f_{\mathrm{MLP}}}

\newcommand{\relu}{\mathrm{RELU}}

\ifcvprfinal\fi
\begin{document}

\title{Composing Text and Image for Image Retrieval - An Empirical Odyssey}
\author{Nam Vo$^1$\thanks{\scriptsize Work done during an internship at Google AI.}, Lu Jiang$^2$, Chen Sun$^2$, Kevin Murphy$^2$ \\
Li-Jia Li$^{2,3}$, Li Fei-Fei$^{2,3}$, James Hays$^1$ \\
$^1$Georgia Tech, $^2$Google AI, $^3$Stanford University \\
}

\maketitle

\begin{abstract}
\eat{
In this paper, we study image retrieval where the query is composed of a reference image and a modification text. Our goal is to find images that are visually similar to the reference image but contain changes stated in the text. We study this problem via a compositional learning perspective and are interested in learning cross-modal composition between images and texts for retrieval. We adapt a standard image retrieval pipeline, investigate existing composition mechanisms, and propose a new composition method of modifying feature via a residual link. Experimental results show that our method performs favorably against existing composition methods, achieving the state-of-the-art result on two datasets. Besides, we also create a simple dataset to motivate the task, establishing a new benchmark for future research on this topic.
}
In this paper, we study the task of image retrieval, where the input query is specified
in the form of an image plus some text that describes desired modifications to the input image.
For example, we may present an image of the Eiffel tower, and ask the system to find
images which are visually similar, but are modified in small ways, such as being taken at nighttime instead of during the day.
To tackle this task, we learn a similarity metric between a target image $x_j$ and a source image $x_i$ plus source text $t_i$,
i.e., a function $\fsim(q_i, x_j)$, where $q_i=\fquery(x_i, t_i)$  is some representation of the query,
such that the similarity is high iff $x_j$ is a ``positive match'' to $q_i$.
We propose a new way to combine image and text using  $\fquery$, that is designed for the retrieval task.
We show this outperforms existing approaches on 3 different datasets,
namely Fashion-200k, MIT-States and a new synthetic dataset we create based on CLEVR.
We also show that our approach can be used to perform image classification
with compositionally novel labels, and we outperform previous methods on MIT-States on this task.
\end{abstract}
\section{Introduction}
\label{sec:intro}

A core problem in image retrieval is that the user has a ``concept'' in mind,
which they want to find images of, but they need to somehow convey that concept
to the system. There are several ways of formulating the concept as a search query,
such as a text string, a similar image, or even a sketch, or some combination of the above.
In this work, we consider the case where queries are formulated as an input image
plus a text string that describes some desired modification to the image.
This represents a typical scenario in session search: users can use an already found image as a reference, 
and then express the difference in text, with the aim of retrieving a relevant image. 
This problem is closely related to attribute-based product retrieval
(see e.g.,  \cite{han2017automatic}), but differs in that the text can be multi-word,
rather than a single attribute.


\begin{figure}[t]
\centering
\includegraphics[width=0.99\linewidth]{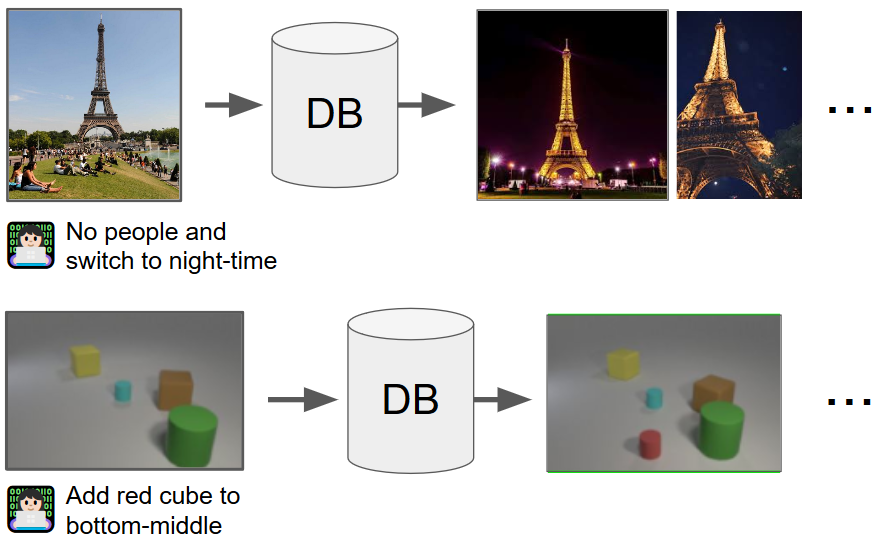}
\caption{Example of image retrieval using text and image query. The text states the desired modification to the image and is expressive in conveying the information need to the system.}
\label{fig:task}
\end{figure}

We can use standard deep metric learning methods such as triplet loss
(e.g., \cite{hermans2017defense})
for computing similarity between a search query and candidate images. The main research question we study is how to represent the query when we have two different input modalities, namely the input image and the text. In other words, how to learn a meaningful cross-modal feature composition for the query in order to find the target image.

Feature composition between text and image has been extensively studied in the field of vision and language, especially in Visual Question Answering (VQA)~\cite{vqa}. After encoding an image (e.g., using a convolutional neural network, or CNN) and the text  (e.g., using a recurrent neural network, or RNN),
various methods for feature composition have been used.
These range from simple techniques (e.g., concatenation or shallow feed-forward networks) to advanced mechanisms (e.g., relation~\cite{santoro2017simple}, or parameter hashing~\cite{noh2016image}). 
These approaches have also been successfully used in related problems such as query classification, compositional learning, etc. (See Section~\ref{sec:related} for more discussion of related work.)

The question of which image/text feature composition to use for image retrieval
has not been studied, to the best of our knowledge.
In this paper, we compare several existing methods, and propose a new one,
which often gives improved results. The key idea behind the new method
is that the text should modify the features of the query image,
but we want the resulting feature vector to still "live in" the same space as the target image.
We achieve this goal by having the text modify the image feature via a gated residual connection.
We call this "\otherfancynameLong" (or \otherfancyname for short).
We give the details in Section~\ref{sec:method}.

We empirically compare these methods on three benchmarks:
Fashion-200k dataset from \cite{han2017automatic},
MIT-States dataset~\cite{StatesAndTransformations},
and a new synthetic dataset for image retrieval, which we call ``CSS'' (color, shape and size), based on the CLEVR framework \cite{johnson2017clevr}.
We show that our proposed feature combination method outperforms existing methods in all three cases.
In particular, significant improvement is made on Fashion-200k compared to \cite{han2017automatic} whose approach is not ideal for this image retrieval task. Besides, our method works reasonably well on a recent task of learning feature composition for image classification~\cite{misra2017red,nagarajan2018attributes}, and achieves the state-of-the-art result on the task on the MIT-States dataset~\cite{StatesAndTransformations}.

To summarize, our contribution is threefold:
\begin{itemize}[leftmargin=*]
\item We systematically study feature composition for image retrieval, and propose a new method.
\item We create a new dataset, CSS, which we will release, which enables controlled experiments of image retrieval using text and image queries.
\item We improve previous state of the art results for image retrieval and compositional image classification on two public benchmarks, Fashion-200K and MIT-States.
\end{itemize}

\section{Related work}
\label{sec:related}

\textbf{Image retrieval and product search}:
Image retrieval is an important vision problem and significant progress has been made thanks to deep learning \cite{chopra2005learning, wang2014learning,gordo2016deep,radenovic2016cnn}; it has numerous applications such as product search \cite{liu2016deepfashion}, face recognition \cite{schroff2015facenet,parkhi2015deep} or image geolocalization \cite{hays2008im2gps}. Cross-modal image retrieval allows using other types of query, examples include text to image retrieval \cite{wang2016learning}, sketch to image retrieval \cite{sangkloy2016sketchy} or cross view image retrieval \cite{lin2015learning}, and event detection \cite{jiang2015bridging}. We consider our set up an image to image retrieval task, but the image query is augmented with an additional modification text input.

A lot of research has been done to improve product retrieval performance by incorporating user's feedback to the search query in the form of relevance \cite{rui1998relevance,jiang2012leveraging}, relative \cite{kovashka2012whittlesearch} or absolute attribute \cite{zhao2017memory, han2017automatic, ak2018learning}. Tackling the problem of image based fashion search, Zhao \etal \cite{zhao2017memory} proposed a memory-augmented deep learning system that can perform attribute manipulation. In \cite{han2017automatic}, spatially-aware attributes are automatically learned from product description labels and used to facilitate attribute-feedback product retrieval application. We are approaching the same image search task, but incorporating text into the query instead, which can be potentially more flexible than using a predefined set of attribute values. Besides, unlike previous work which seldom shows its effectiveness beyond image retrieval, we show our method also work reasonably well for a classification task on compositional learning. 

Parallel to our work is dialog-based interactive image retrieval \cite{guo2018dialog}, where Guo \etal showed promising result on simulated user and real world user study.
Though the task is similar, their study focuses on modeling the interaction between user and the agent; meanwhile we study and benchmark different image text composition mechanisms.

\textbf{Vision question answering}:
The task of Visual Question Answering (VQA) has achieved much attention (see e.g., \cite{vqa, johnson2017clevr}).
Many techniques have been proposed to combine the text and image inputs effectively~\cite{noh2016image, perez2017film,santoro2017simple,liang2018focal}. Generally, these methods aim at learning a ``brand new'' feature which could be difficult in image retrieval. Below we review two closely related work. In \cite{noh2016image}, the text feature is incorporated by mapping into parameters of a fully connected layer within the image CNN. Feature modulation FiLM~\cite{perez2017film} injects text features into image CNN. Though it appears to be similar to our method, they have a few major differences: 1) our modification is learned by the image and text feature together, instead of the text feature alone. 2) the specific modification operations are different. Ours is nonlinear with much more learnable parameters,  versus a linear transformation with few parameters. That is why a FiLM layer is only able to handle a simple operation like scaling, negating or thresholding. 3) As a result, FiLM has to be injected to every layer to perform complex operations while ours is only applied to a single layer. This is essential as modifying as few layer as possible helps ensure the modified feature lives in the same space of the target image.

\textbf{Compositional Learning}:
We can think of our query as a composition of an image and a text. The core of compositional learning is that a complex concept can be developed by combing multiple simple concepts or attributes~\cite{misra2017red}. The idea is reminiscent of earlier work on visual attribute~\cite{farhadi2009describing,sadeghi2011recognition} and also related to zero-shot learning~\cite{lampert2009learning,romera2015embarrassingly,zhang2015zero}. Among recent contributions, Misra \etal~\cite{misra2017red} investigated learning a composition classifier by combining an existing object classifier and attribute classifier. Nagarajan \etal \cite{nagarajan2018attributes} proposed an embedding approach to carry out the composition using the attribute embedding as an operator to change the object classifier. Kota \etal~\cite{kato2018compositional} applied this idea to action recognition. By contrast, our composition is cross-modal and only has a single image versus abundant training examples to train the classifiers.


\section{Method}
\label{sec:method}

\begin{figure*}[ht]
\centering
\includegraphics[width=0.99\linewidth, ]{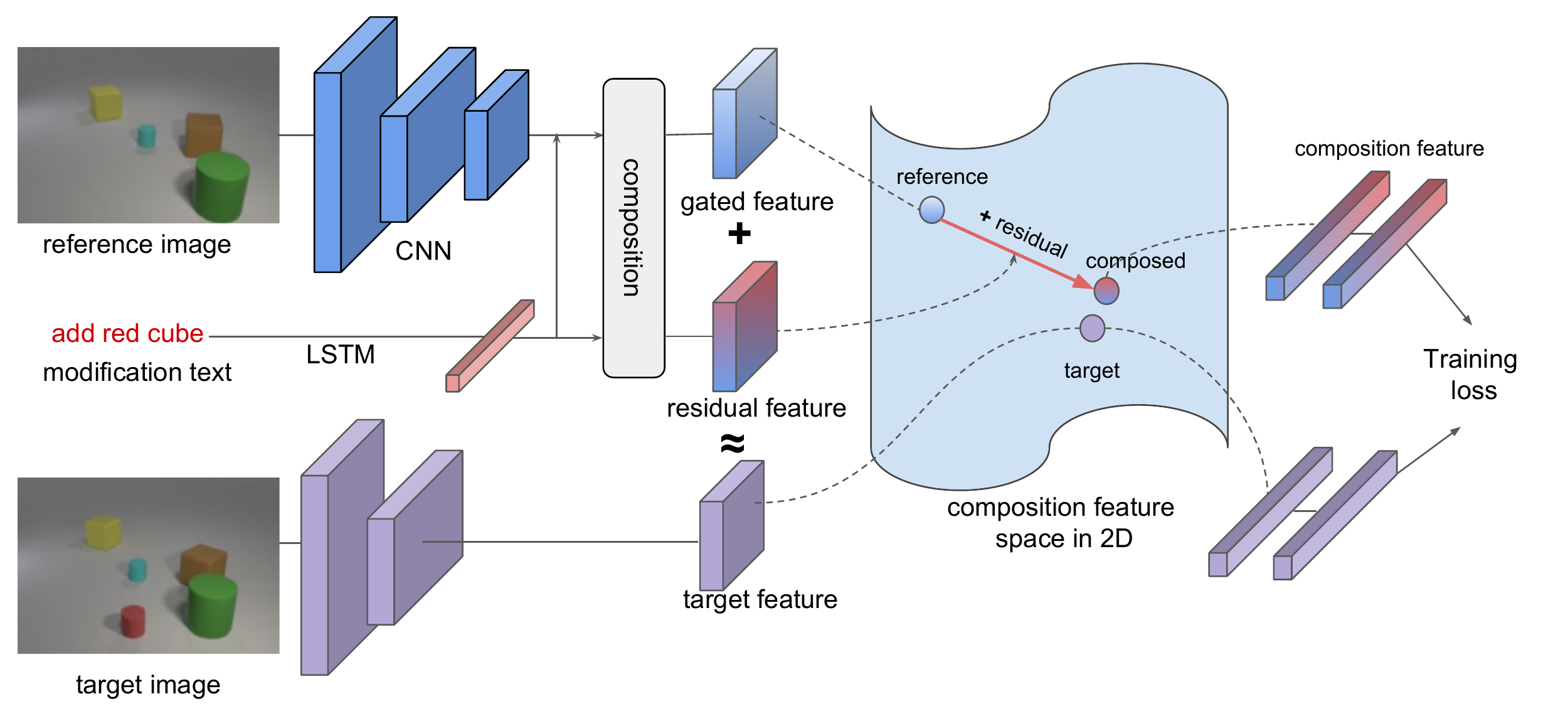}
\caption{The system pipeline for training. We show a 2d feature space for visual simplicity.}
\label{fig:modviares}
\vspace{-4mm}
\end{figure*}

As explained in the introduction, our goal is to learn an embedding space for the text+image query and for target images,
such that matching (query, image) pairs are close
(see  Fig.~\ref{fig:modviares}).

First, we encode the query (or reference) image $x$ using a ResNet-17 CNN to get a 2d spatial feature vector $\fimg(x) = \phi_x \in \mathbb{R}^{W \times H \times C}$,
where $W$ is the width, $H$ is the height, and $C=512$ is the number of feature channels. Next we encode the query text $t$ using a standard LSTM. We define $ \ftext(t)=\phi_t \in \mathbb{R}^d$ to be the hidden state at the final time step whose size $d$ is 512. We want to keep the text encoder as simple as possible. Encoding texts by other encoders, e.g. bi-LSTM or LSTM attention, is definitely feasible but beyond the scope of our paper.
Finally, we combine the two features to compute
$\phi_{xt} = \fquery(\phi_x,\phi_t)$. Below we discuss various ways to perform this combination.

\subsection{Summary of existing combination methods}

In this paper, we study the following approaches for feature composition. For a fair comparison, we train all methods including ours using the same pipeline, with the only difference being in the composition module.
\begin{itemize}[leftmargin=*]

\vspace{-1mm}

\item Image Only: we set $\phi_{xt}=\phi_x$.

\vspace{-1mm}

\item Text Only: we set $\phi_{xt}=\phi_t$.

\vspace{-1mm}

\item Concatenate computes $\phi_{xt}= \MLP([\phi_x,  \phi_t])$. This simple has proven effective in a variety of applications~\cite{vqa,guo2018dialog,zhao2017memory,misra2017red}. In particular, we use two layers of MLP with $\relu$, the batch-norm and the dropout rate of $0.1$.

\vspace{-1mm}

\item Show and Tell~\cite{vinyals2015show}. In this approach, we train a LSTM to encode both image and text by inputting the image feature first, following by words in the text; the final state of this LSTM is used as representation $\phi_{xt}$.

\vspace{-1mm}

\item Attribute as Operator~\cite{nagarajan2018attributes} embeds each text as a transformation matrix, $T_t$, and applies $T_t$ to $\phi_x$ to create $\phi_{xt}$.

\vspace{-1mm}

\item Parameter hashing~\cite{noh2016image} is a technique used for the VQA task. In our implementation, the encoded text feature $\phi_t$ is hashed into a transformation matrix $T_t$, which can be applied to image feature; it is used to replace a fc layer in the image CNN, which now outputs a representation $\phi_{xt}$ that takes into account both image and text feature.

\vspace{-1mm}

\item Relationship~\cite{santoro2017simple} is a method to capture relational reasoning in the VQA task. It first uses CNN to extract a 2d feature map from image, then create a set of relationship features, each is a concatenation of the text feature  $\phi_{t}$ and 2 local features in the 2d feature map; this set of features is passed through a MLP and the result is averaged to get a single feature  $\phi_{xt}$. 

\vspace{-1mm}
\item FiLM~\cite{perez2017film} is another VQA method where the text feature is also injected into the image CNN.
In more detail, the text feature $\phi_t$ is used to predict modulation features: $\gamma^i, \beta^i \in \mathbb{R}^C$, where $i$ indexes the layer and $C$ is the number of feature or feature map. Then it performs a feature-wise affine transformation of the image features,
$\phi^i_{xt} = \gamma^i \cdot \phi^i_x + \beta^i$. As stated in~\cite{perez2017film}, a FiLM layer only handles a simple operation like scaling, negating or thresholding the feature. To perform complex operations, it has to be used in every layer of the CNN.
By contrast, we only modify one layer  of the image feature map,
and we do this using a gated residual connection, described in \ref{sec:ourcombo}.


\end{itemize}


\subsection{Proposed approach: \otherfancyname}
\label{sec:ourcombo}


We propose to combine  image and text features using the following approach
which we call \otherfancynameLong (or \otherfancyname for short).\footnote{
It is possible that an end-to-end learning approach could discover this decomposition automatically; however, manually adding in this form of inductive bias can help reduce the sample complexity, which is an important issue since it is difficult to obtain large paired image-text datasets.
}
\begin{align}
\label{equation:morel}
\phi_{xt}^{rg}
= w_g \fgate (\phi_x, \phi_t) + w_r  \fres (\phi_x, \phi_t),
\end{align}
where $\fgate, \fres \in \mathbb{R}^{W \times H \times C}$ are the gating and the residual features shown in Fig.~\eqref{fig:modviares}. $w_g, w_r$ are learnable weights to balance them. 
The gating connection is computed by:
\begin{align}
\fgate(\phi_x,\phi_t) \!=\!\sigma(W_{g2} \!*\! \relu(W_{g1} \!*\![\phi_x, \! \phi_t]) \odot \phi_x
\end{align}
where $\sigma$ is the sigmoid function, $\odot$ is element wise product,
 $*$ represents 2d convolution with batch normalization,
and $W_{g1}$ and $W_{g2}$ are 3x3 convolution filters.
Note that we broadcast $\phi_t$ along the height and width dimension so that its shape is compatible to the image feature map $\phi_x$.
The residual connection is computed by:
\begin{align}
\fres(\phi_x, \phi_t) = W_{r2} * \relu(W_{r1} * ([\phi_x, \phi_t])),
\end{align}
Eq.~\eqref{equation:morel} combines these two feature vectors using addition. The intuition is that we just want to ``modify'' the image feature based on the text feature, rather than create an entirely different feature space. The gating connection is designed to retain the query image feature, which is helpful if the text modification is insignificant, e.g., with respect to only a certain region in the image. 

Fig.~\ref{fig:modviares} shows modification applied to the convolutional
layer of the CNN. 
However, we can alternatively
apply modification to  the fully-connected layer (where $W=H=1$) to alter the non-spatial properties of the representation. In our experiments, we modify the last fc layer for Fashion200k and MIT-States, since the modification is more global and abstract. For CSS, we modify the last 2d feature map before pooling (last conv layer) to capture the low-level and spatial changes inside the image. 
The choice of which layer to modify is a hyperparameter of the method and can be chosen based on a validation set.

\subsection{Deep Metric Learning}
\label{sec:DML}
\newcommand{\query}{\psi_i}

Our training objective is to push closer the features of the ``modified'' and target image, while pulling apart the features of non-similar images. We employ a classification loss for this task.
More precisely, suppose we have a training minibatch of $B$ queries,
where $\query=\fquery(x^{query}_i, t_i)$ is the final modified representation (from the last layer) of the image text query, and $\phi^+_i=\fimg(x^{target}_i)$ is the representation of the target image of that query.
We create a set $\mathcal{N}_i$ consisting of one positive example $\phi_i^+$ and $K-1$ negative examples $\phi_{1}^-,\ldots, \phi_{K-1}^-$ (by sampling from the minibatch $\phi^+_j$ where $j$ is not $i$). We repeat this $M$ times, denoted as $\mathcal{N}_i^m$, to evaluate every possible set.
(The maximum value of $M$ is $\binom B K$,
but we often use a smaller value for tractability.)
We then use the following softmax cross-entropy loss:

\begin{align}
\label{eq:loss_softmax}
L =  \frac{-1}{MB} \sum_{i=1}^B \sum_{m=1}^M \log\{\frac{\exp\{\kappa(\query, \phi^+_i)\}}
{\sum_{\phi_j \in \mathcal{N}_i^m} \exp\{\kappa(\query, \phi_j)\}}\},
\end{align}
where $\kappa$ is a similarity kernel and is implemented as the dot product or the negative $l_2$ distance in our experiments.
When we use the smallest value of $K=2$, Eq.~\eqref{eq:loss_softmax} can be easily rewritten as:
\begin{align}
L \!=\!  \frac{1}{MB} \sum_{i=1}^B \sum_{m=1}^M \log\{1 \!+\! \exp\{\kappa(\query,\! \phi_{i,m}^{-}) \!-\! \kappa(\query,\! \phi^+_i)\}\},
\end{align}
since each set $\mathcal{N}_i^m$ contains a single negative example.
This is equivalent to the the soft triplet based loss used in \cite{vo2016localizing,hermans2017defense}.
When we use $K=2$, we choose $M=B-1$, so we pair each example $i$ with all possible negatives.

If we use larger $K$, each example is contrasted with a set of other negatives;
this loss resembles the classification based loss used in \cite{goldberger2005neighbourhood,movshovitz2017no,snell2017prototypical,gidaris2018dynamic}.
With the largest value $K=B$, we have $M=1$, so the function is simplified as:
\begin{align}
\label{eq:loss_softmax}
L =  \frac{1}{B}  \sum_{i=1}^B - \log\{\frac{\exp\{\kappa(\query, \phi^+_i)\}}
{\sum_{j=1}^B \exp\{\kappa(\query, \phi^+_j)\}}\},
\end{align}
In our experience, this case is more discriminative and fits faster, but can be more vulnerable to overfitting. As a result, we set $K=B$ for Fashion200k since it is more difficult to converge and $K=2$ for other datasets. Ablation studies on $K$ are shown in Table~\ref{tab:ablation}.

\section{Experiments}

We perform our empirical study on three datasets: Fashion200k~\cite{han2017automatic}, MIT-States~\cite{StatesAndTransformations}, and a new synthetic dataset we created called CSS (see Section~\ref{sec:css}).
Our main metric for retrieval is recall at rank k (R@K), computed as the percentage of test queries where (at least 1) target or correct labeled image is within the top K retrieved images. Each experiment is repeated 5 times to obtain a stable retrieval performance, and both mean and standard deviation are reported. In the case of MIT-States, we also report classification results.

We use PyTorch in our experiments. We use ResNet-17 (output feature size = 512) pretrained on ImageNet as our image encoder and the LSTM (hidden size is 512) of random initial weights as our text encoder. By default, training is run for 150k iterations with a start learning rate 0.01. We will release the code and CSS dataset to the public. Using the same training pipeline, we implement and compare various methods for combining image and text, described in section 3.1, with our feature modification via residual values, described in section 3.2, denoted as \textbf{\otherfancyname}.


\subsection{Fashion200k}

Fashion200k~\cite{han2017automatic} 
is a challenging dataset consisting of $\sim$200k images of fashion products. Each image comes with a compact attribute-like product description (such as black biker jacket or wide leg culottes trouser).
Following~\cite{han2017automatic}, queries are created as following: pairs of products that have one word difference in their descriptions are selected as the query images and target images; and the modification text is that one different word. We used the same training split (around 172k images) and generate queries on the fly for training. To compare with~\cite{han2017automatic}, we randomly sample 10 validation sets of 3,167 test queries (hence in total 31,670 test queries) and report the mean.\footnote{
We contacted the authors of~\cite{han2017automatic} for the original 3,167 test queries, but got only the product descriptions. We attempted to recover the set from the description. However, on average, there are about 3 product images for each unique product description.}.



\begin{figure*}
\centering
\includegraphics[width=0.99\linewidth]{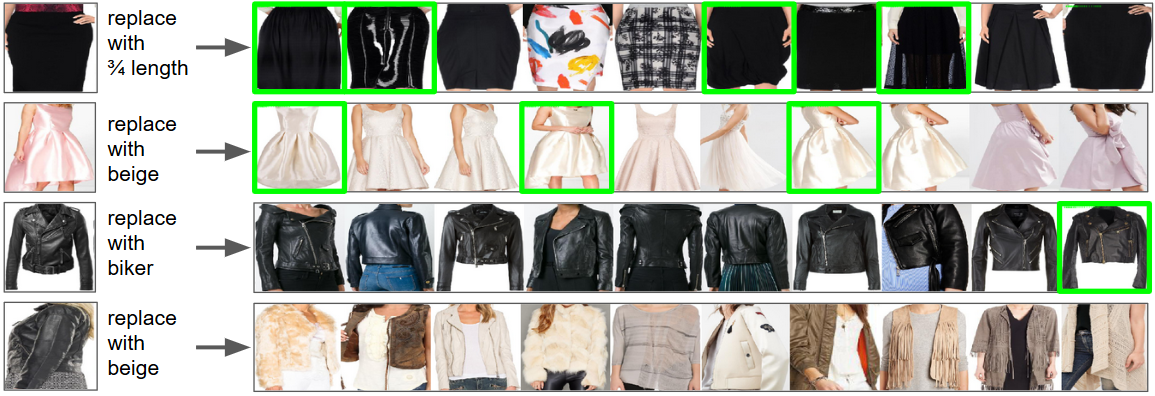}
\caption{Retrieval examples on Fashion200k dataset.}
\label{fig:f200kqual}
\end{figure*}

Table~\ref{tab:f200k} shows the results, where the recall of the first row is from~\cite{han2017automatic} and the others are from our framework. We
see that all of our methods outperform their approach.
We believe this is because they force images and text to be embedded into the same space,
rather than using the text to modify the image space.
In terms of the different ways of computing $\phi_{xt}$,
we see that our approach performs the best.
Some qualitative retrieval examples are shown in Fig.~\ref{fig:f200kqual}.

\begin{table}[ht]
\centering
\begin{tabular}{|l|ccc|}
\hline
Method & R@1 & R@10 & R@50 \\
\hline\hline
Han \etal~\cite{han2017automatic} & 6.3 & 19.9 & 38.3 \\
\hline
Image only & 3.5 & 22.7 & 43.7 \\
Text only & 1.0 & 12.3 & 21.8 \\
Concatenation & 11.9$^{\pm1.0}$ & 39.7$^{\pm1.0}$ & \underline{62.6}$^{\pm0.7}$ \\
Show and Tell & 12.3$^{\pm1.1}$ & 40.2$^{\pm1.7}$ & 61.8$^{\pm0.9}$ \\
Param Hashing & 12.2$^{\pm1.1}$ & 40.0$^{\pm1.1}$ & 61.7$^{\pm0.8}$ \\
Relationship & \underline{13.0}$^{\pm0.6}$ & \underline{40.5}$^{\pm0.7}$ & 62.4$^{\pm0.6}$ \\
FiLM  & 12.9$^{\pm0.7}$ & 39.5$^{\pm2.1}$ & 61.9$^{\pm1.9}$ \\
\otherfancyname & \textbf{14.1}$^{\pm0.6}$ & \textbf{42.5}$^{\pm0.7}$ & \textbf{63.8}$^{\pm0.8}$ \\
\hline
\end{tabular}
\vspace{2mm}
\caption{Retrieval performance on Fashion200k. The best number is in bold and the second best is underlined.}
\label{tab:f200k}
\vspace{-4mm}
\end{table}

\subsection{MIT-States}

MIT-States~\cite{StatesAndTransformations} has $\sim$60k images, each comes with an object/noun label and a state/adjective label (such as ``red tomato'' or ``new camera''). There are 245 nouns and 115 adjectives, on average each noun is only modified by $\sim$9 adjectives it affords.  We use it to evaluate both image retrieval and image classification, as we explain below.

\subsubsection{Image retrieval}

We use this dataset for image retrieval as follows: pairs of images with the same object labels and different state labeled are sampled. They are using as query image and target image respectably. The modification text will be the state of the target image. Hence the system is supposed to retrieve images of the same object as the query image, but with the new state described by text. We use 80 of the nouns for training, and the rest is for testing. This allows the model to learn about state/adjective (modification text) during training and has to deal with unseen objects presented in the test query.

\begin{figure*}
\begin{center}
\includegraphics[width=0.99\linewidth]{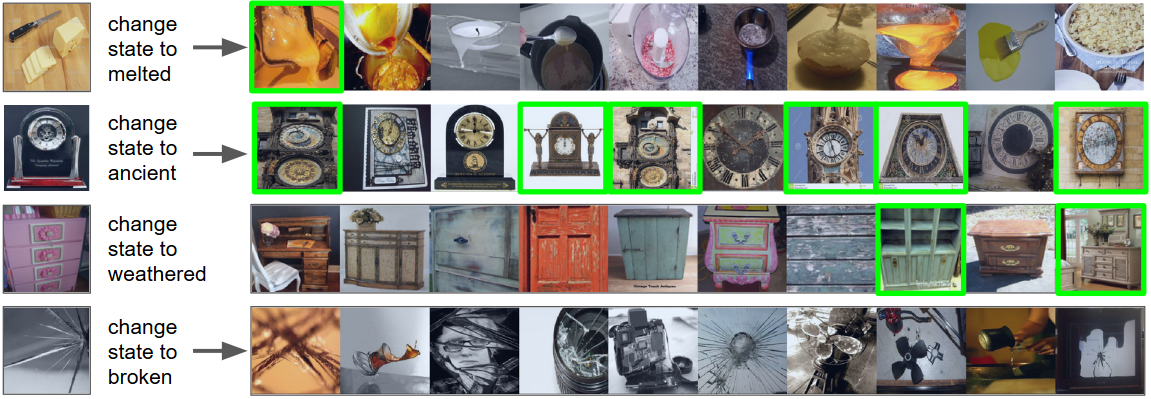}
\end{center}
   \caption{Some retrieval examples on MIT-States.}
\vspace{2mm}
\label{fig:mitstatesqual}
\vspace{-3mm}
\end{figure*}

\begin{table}
\centering
\begin{tabular}{|l|ccc|}
\hline
Method & R@1 & R@5 & R@10 \\
\hline\hline
Image only & 3.3$^{\pm0.1}$ & 12.8$^{\pm0.2}$ & 20.9$^{\pm0.1}$ \\
Text only & 7.4$^{\pm0.4}$ & 21.5$^{\pm0.9}$ & 32.7$^{\pm0.8}$ \\
Concatenation & 11.8$^{\pm0.2}$ & 30.8$^{\pm0.2}$ & 42.1$^{\pm0.3}$ \\
Show and Tell & 11.9$^{\pm0.1}$ & 31.0$^{\pm0.5}$ & 42.0$^{\pm0.8}$ \\
Att. as Operator  & 8.8$^{\pm0.1}$ & 27.3$^{\pm0.3}$ & 39.1$^{\pm0.3}$ \\
Relationship  & \textbf{12.3}$^{\pm0.5}$ & \textbf{31.9}$^{\pm0.7}$ & \underline{42.9}$^{\pm0.9}$ \\
FiLM  & 10.1$^{\pm0.3}$ & 27.7$^{\pm0.7}$ & 38.3$^{\pm0.7}$ \\
\otherfancyname & \underline{12.2}$^{\pm0.4}$ & \textbf{31.9}$^{\pm0.3}$ & \textbf{43.1}$^{\pm0.3}$ \\
\hline
\end{tabular}
\vspace{2mm}
\caption{Retrieval performance on MIT-States.}
\label{tab:mitstates}
\vspace{-3mm}
\end{table}

Some qualitative results are shown in Fig.~\ref{fig:mitstatesqual} and the R@K performance is shown in Table~\ref{tab:mitstates}. Note that similar types of objects with different states can look drastically different, making the the role of modification text more important. Hence on this dataset, the "Text only" baseline outperforms "Image only".
Nevertheless, combining them gives better results. The performance of the exact combination mechanism appears to be less significant on this dataset.
TIRG and Relationship are comparable while outperforming other composition methods.

\subsubsection{Classification with compositionally novel labels}

To be able to compare to prior work on this dataset,
we also consider the classification setting proposed in \cite{misra2017red,nagarajan2018attributes}.
The goal is to learn models to recognize unseen combination of (state, noun) pairs. For example, training on ``red wine'' and ``old tomato'' to recognize ``red tomato'' where there exist no ``red tomato'' images in training.

To tackle this in our framework, we define $\phi_x$ to be the feature vector derived from the image $x$ (using ResNet-17 as before),
and $\phi_t$ to be the feature vector derived from the text $t$.
The text is composed of two words, a noun $n$ and an adjective $a$.
We learn an embedding for each of these words, $\phi_n$ and $\phi_a$,
and then use our \otherfancyname method to compute the combination $\phi_{an}$.
Given this, we perform image classification using nearest neighbor retrieval,
so $t(x) = \arg \max_t \kappa(\phi_t, \phi_x)$,
where $\kappa$ is a similarity kernel applied to the learned embeddings.
(In contrast to our other experiments, here we embed text and image into the same shared space.)

The results, using the same compositional split as in \cite{misra2017red,nagarajan2018attributes},
are shown in Table~\ref{tab:mitstatesclassification}. 
Even though this problem is not the focus of our study, we see that our method outperforms prior methods on this task.
The difference 
from the previous best method, \cite{nagarajan2018attributes}, 
is that their composition feature is represented as a dot product between adjective transformation matrix and noun feature vector; by contrast, we represent both adjective and noun as feature vectors and combine them using our composition mechanism.

\begin{table}
\centering
\begin{tabular}{|l|c|}
\hline
Method & Accuracy \\
\hline\hline
Analogous Attribute \cite{chen2014inferring} & 1.4 \\
Red wine \cite{misra2017red} & 13.1 \\
Attribute as Operator \cite{nagarajan2018attributes}  & 14.2 \\
VisProd NN \cite{nagarajan2018attributes} & 13.9 \\
Label Embedded+ \cite{nagarajan2018attributes} & 14.8 \\
\otherfancyname & \textbf{15.2} \\
\hline
\end{tabular}
\vspace{2mm}
\caption{Comparison to the state-of-the-art on the unseen combination classification task on MIT-States. All baseline numbers are from previous works.}
\label{tab:mitstatesclassification}
\vspace{-3mm}
\end{table}

\subsection{CSS dataset}\label{sec:css}

Since existing benchmarks for image retrieval do not contain complex text modifications, we create a new dataset, as we describe below.

\subsubsection{Dataset Description}

We created a new dataset using the CLEVR toolkit~\cite{johnson2017clevr} for generating synthesized images in a 3-by-3 grid scene. We render objects with different \emph{C}olor, \emph{S}hape and \emph{S}ize (CSS) occupy. Each image comes in a simple 2D blobs version and a 3D rendered version. Examples are shown in Fig.~\ref{fig:cssqual}. 

\lu{check the rest of Sec. 4.3.1}
We generate three types of modification texts from templates: adding, removing or changing object attributes. The ``add object'' modification specifies a new object to be placed in the scene (its color, size, shape, position). If any of the attribute is not specified, its value will be randomly chosen. Examples are ``add object'', ``add red cube'', ``add big red cube to middle-center''.
Likewise, the ``remove object'' modification specifies the object to be removed from the scene. All objects that match the specified attribute values will be removed, e.g. ``remove yellow sphere'', ``remove middle-center object''. Finally, the ``change object'' modification specifies the object to be changed and its new attribute value. The new attribute value has to be either color or size. All objects that match the specified attribute will be changed, e.g. ``make yellow sphere small'', ``make middle-center object red''.

In total, we generate 16K queries for training and 16K queries for test. Each query is of a reference image (2D or 3D) and a modification, and the target image. To be specific, we first generate 1K random scenes as the reference. Then we randomly generate modifications and apply them to the reference images, resulting in a set of 16K target images. In this way, one reference image can be transformed to multiple different target images, and one modification can be applied to multiple different reference images. We then repeat the process to generate the test images.
We follow the protocol proposed in 
\cite{johnson2017clevr} in which certain object shape and color combinations only appear in training, and not in testing, and vice versa. This provides a stronger test of generalization.


Although the CSS dataset is simple, it allows us to perform controlled experiments, with multi-word text queries,
similar to the CLEVR dataset.
In particular, we can create queries using a 2d image and text string, to simulate the case where the user is sketching something, and then wants to modify it using language.
We can also create queries using slightly more realistic 3d image and text strings.

\eat{
Indeed, this particular retrieval problem is not useful in practice. However, CSS dataset is designed to be a flexible and configurable benchmark for advancing image retrieval, the role of which is similar to the role of VQA for the real-world QA system. CSS allows us to control the task complexity and analyze results by configuring settings of image types (2D or 3D) and modification types (add, remove or modify). For example, we create an interesting 2D-to-3D image retrieval setting. It simulates the scenario where users can express their search query by a simple 2d sketch combined with text modification to be applied on the sketch. The image plus text query is used by our system to retrieval the real 3D images.
}

\begin{figure}[ht]
\centering
\includegraphics[width=0.99\linewidth]{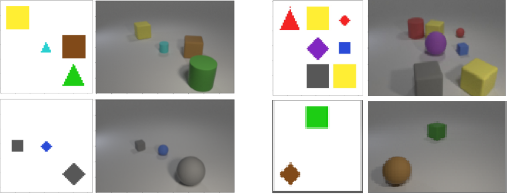}
\caption{Example images in our CSS dataset. The same scene are rendered in 2D and 3D images.}
\label{fig:css_example}
\end{figure}

\subsubsection{Results}

\begin{table}[h]
\centering
\begin{tabular}{|l|c|c|}
\hline
Method & 3D-to-3D & 2D-to-3D \\
\hline\hline
Image only & 6.3 & 6.3 \\
Text only & 0.1 & 0.1 \\
Concatenate & 60.6$^{\pm0.8}$ & 27.3 \\
Show and Tell & 33.0$^{\pm3.2}$ & 6.0 \\
Parameter hashing & 60.5$^{\pm1.9}$ & 31.4 \\
Relationship & 62.1$^{\pm1.2}$ & 30.6 \\
FiLM & \underline{65.6}$^{\pm0.5}$ & \underline{43.7} \\
\otherfancyname &  \textbf{73.7}$^{\pm1.0}$ & \textbf{46.6} \\
\hline
\end{tabular}
\vspace{2mm}
\caption{Retrieval performance (R@1) on the CSS Dataset using 2D and 3D images as the query.}
\label{tab:css}
\vspace{-3mm}
\end{table}

\begin{figure*}[ht]
\centering
\includegraphics[width=0.99\linewidth]{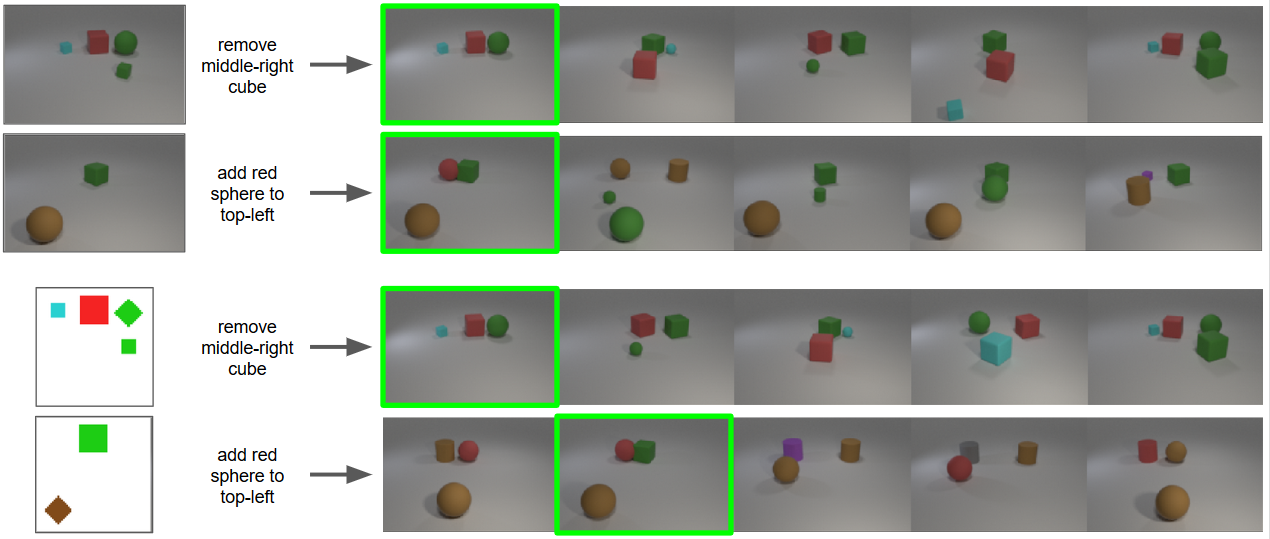}
\caption{Some retrieval examples on CSS Dataset.}
\label{fig:cssqual}
\vspace{-2mm}
\end{figure*}

\eat{
Table~\ref{tab:css} summarizes R@1 retrieval performance on the CSS dataset. We examine two retrieval settings using 3d query images (2nd column) and 2d images (3rd column). As shown, our method performs well against the state-of-the-art methods in both settings, improving the best baseline by a relative 6-12\%. The results show that our method is effective for learning composition feature for image retrieval. 

Fig.~\ref{fig:cssqual} plots some queries and retrieved image by our method. As we see, the 2D-to-3D setting is quite challenging which uses a 2D image and a text to search the database of 3D images. Our improvement on 2D-3D retrieval is smaller. This is probably because the feature of 2d and 3d images are very different, and thus learning new features becomes a reasonable idea. Nevertheless, our method is able to find reasonable results. Notice that we use separate encoders to extract features for 2d and 3d images in this setting.
}

Table~\ref{tab:css} summarizes R@1 retrieval performance on the CSS dataset. We examine two retrieval settings using 3d query images (2nd column) and 2d images (3rd column).
As we can see, our \otherfancyname combination outperforms other composition methods for the retrieval task.
In addition, we see that retrieving a 3D image from a 2D query is much harder, since the feature spaces are quite different. (In these experiments, we use different feature encoders for the 2D and 3D inputs).
Some qualitative results are shown in Fig.~\ref{fig:cssqual}.



\begin{figure*}[ht]
\centering
\includegraphics[width=0.99\linewidth]{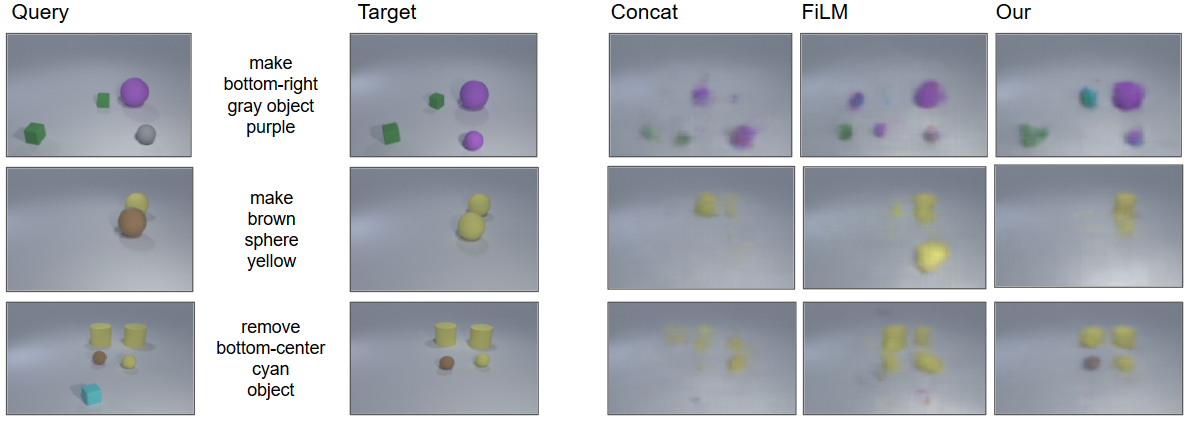}
\caption{Reconstruction images from the learned composition features.}
\label{fig:cssreconstruction}
\vspace{-2mm}
\end{figure*}

To gain more insight into the nature of the combined features, we trained  a transposed convolutional network to reconstruct the images from their features and then apply it to composition feature. Fig.~\ref{fig:cssreconstruction} shows the reconstructed images from the composition features of three methods. 
Images generated from our feature representation look visually better, and are closer to the top retrieved image.
We see that all the images are blurry as we use the regression loss to train the network. 
However,
a nicer reconstruction may not mean better retrieval, as the composition feature is learned to capture the discriminative information need to find the target image, and this may be a lossy representation.

\subsection{Ablation Studies}

\begin{table}[ht]
\centering
\small
\begin{tabular}{|l|ccc|}
\hline
Method     & Fashion & MIT-States & CSS \\
\hline\hline
Our Full Model &   14.1    &       12.2    & 73.7    \\
 - gated feature only  &   13.9      & 07.1   & 06.5  \\
 - residue feature only  &  12.1     &  11.9    & 60.6 \\
 \hline
 - mod. at last fc  &  14.1      &  12.2    & 71.2  \\
 - mod. at last conv    &  12.4    &  10.3   &  73.7 \\
 \hline
 DML loss, $K=2$   &    9.5     &  12.2    & 73.7 \\
 DML loss, $K=B$   &    14.1    &  10.9     & 69.8 \\
\hline
\end{tabular}
\vspace{2mm}
\caption{\label{tab:ablation}Retrieval performance (R@1) of ablation studies.}
\end{table}

In this section, we report the results of various ablation studies,
to gain insight into which parts of our approach matter the most.
The results are in  Table~\ref{tab:ablation}.

\noindent\textbf{Efficacy of feature modification}: as shown in Fig.~\ref{fig:modviares}, our composition module has two types of connections, namely residual connection and gated connection. Row 2 and 3 show that removing the residual
features or gating features leads to drops in performance. In these extreme cases, our model can degenerate to the concatenate fusion baseline.

\noindent\textbf{Spatial versus non-spatial modification}: 
Row 5 and 6 compares the effect of applying our feature modification to the last fc layer versus the last convolution layer. When our modification is applied to the last fc layer feature, it yields competitive performance compared to the baseline across all datasets. Applying the modification to the last convolution feature map only improves the performance on CSS. We believe this is because the modifications in the CSS dataset is more  spatially localized (see Fig.~\ref{fig:cssreconstruction}) whereas they are more global on the other two datasets (See Fig.~\ref{fig:mitstatesqual} and Fig.~\ref{fig:f200kqual})

\noindent\textbf{The impact of $K$ in the loss function}: The last two rows compares the loss function of two different $K$ values in Section~\ref{sec:DML}. We use $K=2$ (soft triplet loss) in most experiments. As Fashion200k is much bigger, we found that the network underfitted. In this case by using $K=B$ (same as batch size in our experiment), the network fits well and produces better results. On the other two datasets, test time performance is comparable, but training becomes less stable. Note that the difference here regards our metric learning loss and does not reflect the difference between the feature composition methods.

\section{Conclusion}
In this work, we explored the composition of image and text in the context of image retrieval.
We experimentally evaluated several existing methods, and proposed a new one,
which gives improved performance on three benchmark datasets.
In the future, we would like to try to scale this method up to work on real image retrieval systems "in the wild".

{\small
\bibliographystyle{ieee}
\bibliography{egbib}

\begin{thebibliography}{10}\itemsep=-1pt

\bibitem{ak2018learning}
K.~E. Ak, A.~A. Kassim, J.~H. Lim, and J.~Y. Tham.
\newblock Learning attribute representations with localization for flexible
  fashion search.
\newblock In {\em CVPR}, 2018.

\bibitem{vqa}
S.~Antol, A.~Agrawal, J.~Lu, M.~Mitchell, D.~Batra, C.~L. Zitnick, and
  D.~Parikh.
\newblock {VQA}: {V}isual {Q}uestion {A}nswering.
\newblock In {\em ICCV}, 2015.

\bibitem{chen2014inferring}
C.-Y. Chen and K.~Grauman.
\newblock Inferring analogous attributes.
\newblock In {\em CVPR}, 2014.

\bibitem{chopra2005learning}
S.~Chopra, R.~Hadsell, and Y.~LeCun.
\newblock Learning a similarity metric discriminatively, with application to
  face verification.
\newblock In {\em CVPR}, 2005.

\bibitem{farhadi2009describing}
A.~Farhadi, I.~Endres, D.~Hoiem, and D.~Forsyth.
\newblock Describing objects by their attributes.
\newblock In {\em CVPR}, 2009.

\bibitem{gidaris2018dynamic}
S.~Gidaris and N.~Komodakis.
\newblock Dynamic few-shot visual learning without forgetting.
\newblock In {\em CVPR}, 2018.

\bibitem{goldberger2005neighbourhood}
J.~Goldberger, G.~E. Hinton, S.~T. Roweis, and R.~R. Salakhutdinov.
\newblock Neighbourhood components analysis.
\newblock In {\em NIPS}, 2005.

\bibitem{gordo2016deep}
A.~Gordo, J.~Almaz{\'a}n, J.~Revaud, and D.~Larlus.
\newblock Deep image retrieval: Learning global representations for image
  search.
\newblock In {\em ECCV}, 2016.

\bibitem{guo2018dialog}
X.~Guo, H.~Wu, Y.~Cheng, S.~Rennie, and R.~S. Feris.
\newblock Dialog-based interactive image retrieval.
\newblock {\em arXiv preprint arXiv:1805.00145}, 2018.

\bibitem{han2017automatic}
X.~Han, Z.~Wu, P.~X. Huang, X.~Zhang, M.~Zhu, Y.~Li, Y.~Zhao, and L.~S. Davis.
\newblock Automatic spatially-aware fashion concept discovery.
\newblock In {\em ICCV}, 2017.

\bibitem{hays2008im2gps}
J.~Hays and A.~A. Efros.
\newblock Im2gps: estimating geographic information from a single image.
\newblock In {\em CVPR}, 2008.

\bibitem{hermans2017defense}
A.~Hermans, L.~Beyer, and B.~Leibe.
\newblock In defense of the triplet loss for person re-identification.
\newblock {\em arXiv preprint arXiv:1703.07737}, 2017.

\bibitem{StatesAndTransformations}
P.~Isola, J.~J. Lim, and E.~H. Adelson.
\newblock Discovering states and transformations in image collections.
\newblock In {\em CVPR}, 2015.

\bibitem{jiang2012leveraging}
L.~Jiang, A.~G. Hauptmann, and G.~Xiang.
\newblock Leveraging high-level and low-level features for multimedia event
  detection.
\newblock In {\em ACM MM}, 2012.

\bibitem{jiang2015bridging}
L.~Jiang, S.-I. Yu, D.~Meng, T.~Mitamura, and A.~G. Hauptmann.
\newblock Bridging the ultimate semantic gap: A semantic search engine for
  internet videos.
\newblock In {\em ICMR}, 2015.

\bibitem{johnson2017clevr}
J.~Johnson, B.~Hariharan, L.~van~der Maaten, L.~Fei-Fei, C.~L. Zitnick, and
  R.~Girshick.
\newblock Clevr: A diagnostic dataset for compositional language and elementary
  visual reasoning.
\newblock In {\em CVPR}, 2017.

\bibitem{kato2018compositional}
K.~Kato, Y.~Li, and A.~Gupta.
\newblock Compositional learning for human object interaction.
\newblock In {\em ECCV}, 2018.

\bibitem{kovashka2012whittlesearch}
A.~Kovashka, D.~Parikh, and K.~Grauman.
\newblock Whittlesearch: Image search with relative attribute feedback.
\newblock In {\em CVPR}, 2012.

\bibitem{lampert2009learning}
C.~H. Lampert, H.~Nickisch, and S.~Harmeling.
\newblock Learning to detect unseen object classes by between-class attribute
  transfer.
\newblock In {\em CVPR}, 2009.

\bibitem{liang2018focal}
J.~Liang, L.~Jiang, L.~Cao, L.-J. Li, and A.~Hauptmann.
\newblock Focal visual-text attention for visual question answering.
\newblock In {\em CVPR}, 2018.

\bibitem{lin2015learning}
T.-Y. Lin, Y.~Cui, S.~Belongie, and J.~Hays.
\newblock Learning deep representations for ground-to-aerial geolocalization.
\newblock In {\em CVPR}, 2015.

\bibitem{liu2016deepfashion}
Z.~Liu, P.~Luo, S.~Qiu, X.~Wang, and X.~Tang.
\newblock Deepfashion: Powering robust clothes recognition and retrieval with
  rich annotations.
\newblock In {\em CVPR}, 2016.

\bibitem{misra2017red}
I.~Misra, A.~Gupta, and M.~Hebert.
\newblock From red wine to red tomato: Composition with context.
\newblock In {\em CVPR}, 2017.

\bibitem{movshovitz2017no}
Y.~Movshovitz-Attias, A.~Toshev, T.~K. Leung, S.~Ioffe, and S.~Singh.
\newblock No fuss distance metric learning using proxies.
\newblock In {\em ICCV}, 2017.

\bibitem{nagarajan2018attributes}
T.~Nagarajan and K.~Grauman.
\newblock Attributes as operators.
\newblock 2018.

\bibitem{noh2016image}
H.~Noh, P.~Hongsuck~Seo, and B.~Han.
\newblock Image question answering using convolutional neural network with
  dynamic parameter prediction.
\newblock In {\em CVPR}, 2016.

\bibitem{parkhi2015deep}
O.~M. Parkhi, A.~Vedaldi, A.~Zisserman, et~al.
\newblock Deep face recognition.
\newblock In {\em BMVC}, 2015.

\bibitem{perez2017film}
E.~Perez, F.~Strub, H.~De~Vries, V.~Dumoulin, and A.~Courville.
\newblock Film: Visual reasoning with a general conditioning layer.
\newblock 2018.

\bibitem{radenovic2016cnn}
F.~Radenovi{\'c}, G.~Tolias, and O.~Chum.
\newblock Cnn image retrieval learns from bow: Unsupervised fine-tuning with
  hard examples.
\newblock In {\em ECCV}, 2016.

\bibitem{romera2015embarrassingly}
B.~Romera-Paredes and P.~Torr.
\newblock An embarrassingly simple approach to zero-shot learning.
\newblock In {\em ICML}, 2015.

\bibitem{rui1998relevance}
Y.~Rui, T.~S. Huang, M.~Ortega, and S.~Mehrotra.
\newblock Relevance feedback: a power tool for interactive content-based image
  retrieval.
\newblock {\em IEEE Transactions on circuits and systems for video technology},
  8(5):644--655, 1998.

\bibitem{sadeghi2011recognition}
M.~A. Sadeghi and A.~Farhadi.
\newblock Recognition using visual phrases.
\newblock In {\em CVPR}, 2011.

\bibitem{sangkloy2016sketchy}
P.~Sangkloy, N.~Burnell, C.~Ham, and J.~Hays.
\newblock The sketchy database: learning to retrieve badly drawn bunnies.
\newblock {\em ACM Transactions on Graphics (TOG)}, 35(4):119, 2016.

\bibitem{santoro2017simple}
A.~Santoro, D.~Raposo, D.~G. Barrett, M.~Malinowski, R.~Pascanu, P.~Battaglia,
  and T.~Lillicrap.
\newblock A simple neural network module for relational reasoning.
\newblock In {\em NIPS}, 2017.

\bibitem{schroff2015facenet}
F.~Schroff, D.~Kalenichenko, and J.~Philbin.
\newblock Facenet: A unified embedding for face recognition and clustering.
\newblock In {\em CVPR}, 2015.

\bibitem{snell2017prototypical}
J.~Snell, K.~Swersky, and R.~Zemel.
\newblock Prototypical networks for few-shot learning.
\newblock In {\em NIPS}, 2017.

\bibitem{vinyals2015show}
O.~Vinyals, A.~Toshev, S.~Bengio, and D.~Erhan.
\newblock Show and tell: A neural image caption generator.
\newblock In {\em CVPR}, 2015.

\bibitem{vo2016localizing}
N.~N. Vo and J.~Hays.
\newblock Localizing and orienting street views using overhead imagery.
\newblock In {\em ECCV}, 2016.

\bibitem{wang2014learning}
J.~Wang, Y.~Song, T.~Leung, C.~Rosenberg, J.~Wang, J.~Philbin, B.~Chen, and
  Y.~Wu.
\newblock Learning fine-grained image similarity with deep ranking.
\newblock In {\em CVPR}, 2014.

\bibitem{wang2016learning}
L.~Wang, Y.~Li, and S.~Lazebnik.
\newblock Learning deep structure-preserving image-text embeddings.
\newblock In {\em CVPR}, 2016.

\bibitem{zhang2015zero}
Z.~Zhang and V.~Saligrama.
\newblock Zero-shot learning via semantic similarity embedding.
\newblock In {\em ICCV}, 2015.

\bibitem{zhao2017memory}
B.~Zhao, J.~Feng, X.~Wu, and S.~Yan.
\newblock Memory-augmented attribute manipulation networks for interactive
  fashion search.
\newblock In {\em CVPR}, 2017.

\end{thebibliography}
}

\newpage
\clearpage
\newpage
\clearpage
\setcounter{equation}{0}
\setcounter{section}{0}
\setcounter{page}{1}


\end{document}